\documentclass[letterpaper]{article}
\usepackage{iccc}
\usepackage{graphicx}
\usepackage{CJKutf8}
\usepackage{times}
\usepackage{helvet}
\usepackage{courier}
\usepackage{caption}
\usepackage{xcolor}
\usepackage{amssymb}
\usepackage{subcaption}
\title{Calliffusion: Chinese Calligraphy Generation and Style Transfer with Diffusion Modeling}

\author{Qisheng Liao,\textsuperscript{1}
Gus Xia,\textsuperscript{1,2}
Zhinuo Wang,\textsuperscript{2}\\
\textsuperscript{1}{Mohamed bin Zayed University of Artificial Intelligence}\\
\textsuperscript{2}{New York University Shanghai}\\
qisheng.liao@mbzuai.ac.ae,
gus.xia@mbuzai.ac.ae, 
zw2375@nyu.edu}

\begin{document} 
\begin{CJK}{UTF8}{gbsn}
\maketitle
\begin{abstract}
\begin{quote}
In this paper, we propose Calliffusion, a system for generating high-quality Chinese calligraphy using diffusion models. Our model architecture is based on DDPM (Denoising Diffusion Probabilistic Models), and it is capable of generating common characters in five different scripts and mimicking the styles of famous calligraphers. Experiments demonstrate that our model can generate calligraphy that is difficult to distinguish from real artworks and that our controls for characters, scripts, and styles are effective. Moreover, we demonstrate one-shot transfer learning, using LoRA (Low-Rank Adaptation) to transfer Chinese calligraphy art styles to unseen characters and even out-of-domain symbols such as English letters and digits.
\end{quote}
\end{abstract}

\section{Introduction}

\begin{figure}[!h]
    \centering
    \includegraphics[width=0.5\textwidth]{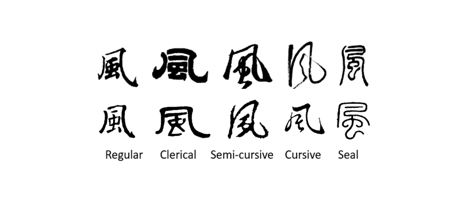}
    \caption{10 samples of the character "风" (wind). }
    \label{fig:feng}
\end{figure}

Chinese calligraphy, which is the artistic writing of Chinese characters and a prominent form of East Asian calligraphy, can be seen as a distinctive form of visual art. There are five Chinese calligraphy scripts, regular (楷), semi-cursive (行), cursive (草), clerical (隶), and seal (篆) script. Regular script is the most common script for writing nowadays. Semi-cursive script is faster to write compared with regular script but is still easily readable. Cursive script is known for its speedy writing style, but it can be challenging to read. Clerical script and seal script nowadays are mainly used for artistic purposes. Besides, each famous calligrapher has his or her own style. Even when they write the same character in the same script, the calligraphy may look very different. For example, Figure \ref{fig:feng} shows 10 samples of the same character. Here, each column belongs to a different script, and for each script, we show two samples of different styles.

Recently, we see a trend in generating Chinese calligraphy using AI, including  Zi2zi\cite{zi2zi}, CalliGAN\cite{wu2020calligan}, and ZiGAN\cite{wen2021zigan}. Most of them adopt a GAN (Generative Adversarial Network)\cite{NIPS2014_5ca3e9b1} architecture, training on paired data of printed font and handwritten font while performing image-to-image translation during inference. Despite some effort on improving the data efficiency \cite{zhou2021end}, two major challenge remains: (1) to generate \textit{high-quality} calligraphy, and (2) to apply effective \textit{controls} on characters, scripts, and calligraphers' styles.

If we visualize the calligraphy artworks in the 3-D space of characters, scripts, and styles, the distribution of the samples would be very \textit{sparse}. There are thousands of Chinese characters, while most calligraphers' work collections only cover a small portion of the characters in particular scripts. In this paper, we aim to model the sample distribution in the 3-D space and generate calligraphy of any character, script, and style. 

To this end, 
we introduce a new method for generating Chinese calligraphy with Denoising Diffusion Probabilistic Models (DDPMs)\cite{ho2020denoising}. In particular, we control the model using external conditions based on Chinese text descriptions of character, script, and style. During training, we utilized labeled calligraphy images, while during inference, we used description texts to control the generation process. Notably, unlike most previous studies that rely on GANs and require an input image for generation, our method does not necessitate the use of any images during inference.

Besides with-in distribution generalization, we also utilize Low-Rank Adaptation of Large Language Models (LoRA)\cite{hu2021lora} to achieve out-of-distribution \textit{style transfer} via one-shot fine-tuning. Experiments show that such an approach can transfer existing scripts and styles to unseen characters and even out-of-domain symbols such as English letters and digits.

Our model could be very useful for individuals engaged in the process of learning Chinese calligraphy. A common approach to learning Chinese calligraphy is to study the artwork of famous calligraphers and imitate their styles. However, the artwork resources of specific calligraphers are usually limited, and it is almost impossible to obtain any character for a specific calligrapher. With the aid of our model, learners can overcome such limitations, generating artworks of any character, any script, and any style.

In summary, the major contributions of our Calliffusion system are:
\begin{itemize}
\item As far as we know, it is the first diffusion model for generating high-quality Chinese calligraphy artwork.
\item The controllable generation is effective. The conditional model can generate calligraphy in any character, script, and calligrapher's style.
\item The style transfer technique is effective. Our one-shot fine-tuning technique can adapt certain scripts and writing styles to unseen Chinese characters and even English letters and digits.
\end{itemize}

\section{Methods}

\subsection{Diffusion Model}
In our research, we used a U-net\cite{ronneberger2015u} model as the backbone model and used DDPMs sampling, which include a forward process (diffusion) that progressively disturbs the organization of the data $x_0$, and a reverse process (denoising) that is trained to restore the initial data $x_0$ from the corrupted input. In this context, $x_0$ refers to the calligraphy image. The forward process involves the addition of Gaussian noise in $N$ diffusion steps as shown in equations \ref{eq:1} and \ref{eq:2} below.

\begin{equation}\label{eq:1}
q(x_{t}|x_{t-1})=\mathcal{N}(x_{t};\sqrt{1-\beta_{t}}x_{t-1},\beta_{t}I) 
\end{equation}

\begin{equation}\label{eq:2}
q(x_{1:T}|x_{0})=\prod_{t=1}^{T}q(x_{t}|x_{t-1})
\end{equation}

The variance scheduling parameters $\beta_1$, $\beta_2$, . . . , $\beta_N$ are employed to regulate the diffusion process. On the other hand, the reverse process requires the model to define a Markov chain that sequentially rebuilds the calligraphy image $x_0$ from a disturbed input $x_N$, which follows a normal distribution $N(0, I)$. The equations \ref{eq:3} and \ref{eq:4} below show the reverse process.

\begin{equation}\label{eq:3}
p_\theta(\mathbf{x}_{0:T}) = p(\mathbf{x}_T) \prod^T_{t=1} p_\theta(\mathbf{x}_{t-1} \vert \mathbf{x}_t)
\end{equation}

\begin{equation}\label{eq:4}
p_\theta(\mathbf{x}_{t-1} \vert \mathbf{x}_t) = \mathcal{N}(\mathbf{x}_{t-1}; \mu_\theta(\mathbf{x}_t, t), \Sigma_\theta(\mathbf{x}_t, t))
\end{equation}
While in the process of training, we aim to minimize the target by optimizing the model parameters represented by $\epsilon_\theta$ as equation \ref{eq:5}, where $t$ is uniformly sampled from $[1, N]$ and $\epsilon \sim \mathcal{N} (0, I)$, $a_t := 1-\beta_t$, $\bar{\alpha}_t:=\prod^t_{s=1}\alpha_s $.

\begin{equation}\label{eq:5}
L(\theta)=\mathbb{E}_{x_0,\epsilon,t} \left [\left \|\epsilon-\epsilon_\theta(\sqrt{\bar{\alpha}_t}x_0+\sqrt{1-\bar{\alpha}_t}\epsilon,t)  \right \|^2 \right]
\end{equation}

\subsection{Adding Controls with External Conditions}
\begin{figure}[h]
    \centering
    \includegraphics[width=0.5\textwidth]{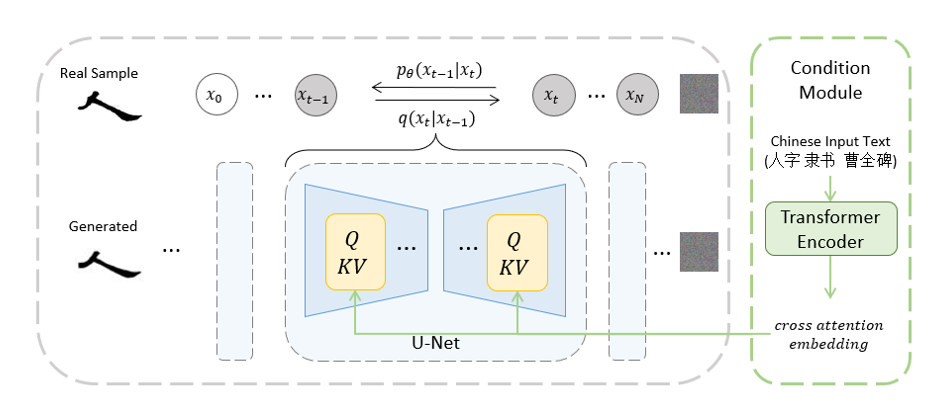}
    \caption{The diffusion model structure with one cross-attention condition that comes from a Transformer encoder.}
    \label{fig:model2}
\end{figure}

In order to control the generations, we rely on three conditions, i.e., characters, scripts, and styles. In specific, We use a short description of Chinese text input, such as '人字 隶书 曹全碑' ('Ren Character, Clerical script, Caoquanbei') to control the generations. The text consists of three parts, and a space separates each part. The first part of the text determines the character, the second part controls the script, and the last part determines the calligrapher's style. The input text is then passed through a pre-trained Chinese BERT model \cite{devlin2018bert} to obtain cross-attention embeddings. These embeddings are combined with the image during the training of the diffusion model. The structure of this conditional model is illustrated in Figure \ref{fig:model2}.

\subsection{Style Transfer via Fine-tuning}
Based on the conditional diffusion model, our Calliffusion system can further transfer the scripts and styles to unseen characters and out-of-domain symbols via one-shot fine-tuning. During the fine-tuning process, we only need to provide the model with a single image of the new character or symbol, either letting its script or style be specified or not specified. After that, the system can generate new calligraphy by applying a script and a style to \textit{that} character or symbol.

The fine-tuning technique is based on LoRA, which is a training technique that speeds up the training process of large models while reducing memory consumption. LoRA achieves this by adding update matrices, which are rank-decomposed weight matrices, to the existing weights, and only updating the newly added weights during training. By keeping the previously pretrained weights frozen, the model is protected against catastrophic forgetting, where it loses previously learned information during further training. Additionally, the rank-decomposition matrices used in LoRA have significantly fewer parameters compared to the original model, making the trained LoRA weights easily transferable and portable.

\begin{figure*}[!htb]
     \centering
     \begin{subfigure}[b]{0.45\textwidth}
         \centering
         \includegraphics[width=\textwidth]{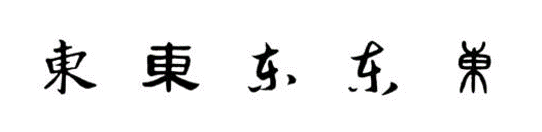}
         \caption{5 generated character 'Dong' in different scripts.}
         \label{fig:dong}
     \end{subfigure}
     \hfill
     \begin{subfigure}[b]{0.45\textwidth}
         \centering
         \includegraphics[width=\textwidth]{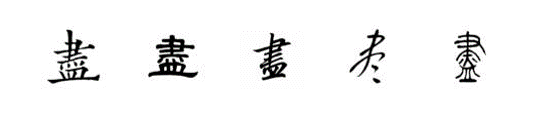}
         \caption{5 generated character 'Jin' in different scripts.}
         \label{fig:lan}
     \end{subfigure}    
     
     \centering
     \begin{subfigure}[b]{0.45\textwidth}
         \centering
         \includegraphics[width=\textwidth]{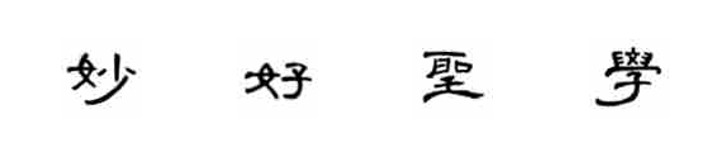}
         \caption{4 characters in clerical script and Caoquanbei's (曹全碑) style. The first image is generated by our model and the real sample for this character in Caoquanbei's style does not exist. The other three images are real Caoquanbei's calligraphy.}
         \label{fig:li_q}
     \end{subfigure}
     \hfill
     \begin{subfigure}[b]{0.45\textwidth}
         \centering
         \includegraphics[width=\textwidth]{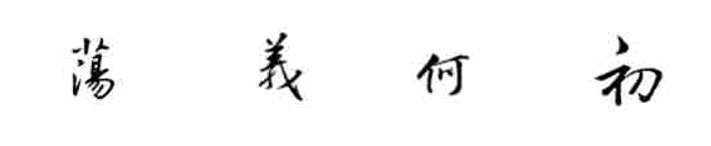}
         \caption{4 characters in semi-cursive script and Wang Xizhi's (王羲之) style. The first image is generated by our model and the real sample for this character in Wang Xizhi's style does not exist. The other three images are real Wang Xizhi's calligraphy.}
         \label{fig:xing_q}
     \end{subfigure}
     \centering
     \begin{subfigure}[b]{0.45\textwidth}
         \centering
         \includegraphics[width=\textwidth]{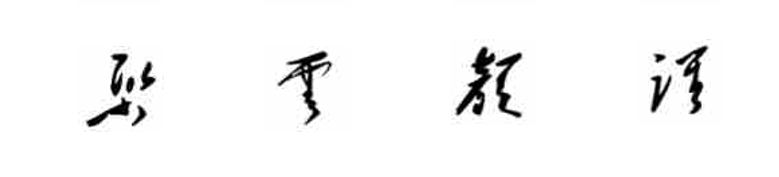}
         \caption{4 characters in cursive script and Mao Zedong's (毛泽东) style. The first image is generated by our model and the real sample for this character in Mao Zedong's style does not exist. The other three images are real Mao Zedong's calligraphy.}
         \label{fig:cao_q}
     \end{subfigure}
     \hfill
     \begin{subfigure}[b]{0.45\textwidth}
         \centering
         \includegraphics[width=\textwidth]{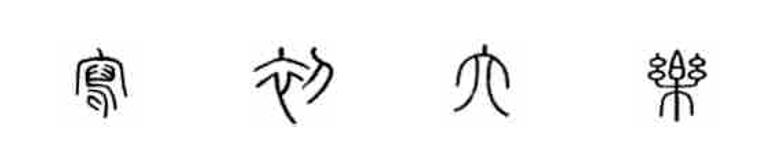}
         \caption{4 characters in seal script and Wang Kuaijishike's (会稽石刻) style. The first image is generated by our model and the real sample for this character in Wang Kuaijishike's style does not exist. The other three images are real Kuaijishike's calligraphy.}
         \label{fig:zhuan_q}
     \end{subfigure}
     \centering
     \begin{subfigure}[b]{0.45\textwidth}
         \centering
         \includegraphics[width=\textwidth]{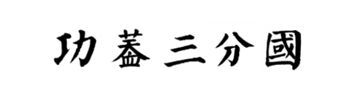}
         \caption{Generated calligraphy for a sentence of a poem. The conditions are regular script and Yan Zhenqin(颜真卿).}
         \label{fig:ggsfg}
     \end{subfigure}
     \hfill
     \begin{subfigure}[b]{0.45\textwidth}
         \centering
         \includegraphics[width=\textwidth]{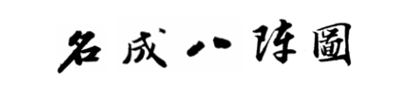}
         \caption{Generated calligraphy for a sentence of a poem. The conditions are semi-cursive script and Su Shi(苏轼).}
         \label{fig:mcbzt}
     \end{subfigure}
     \begin{subfigure}[b]{1\textwidth}
         \centering
         \includegraphics[width=\textwidth]{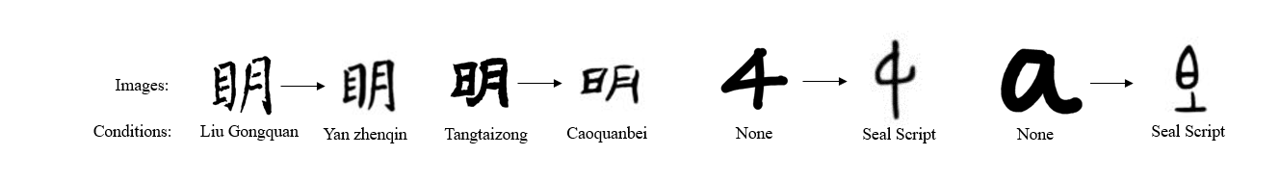}
         \caption{Generations based on with one-shot fine-tuning. The conditions for generation are different from the conditions in fine-tuning but the generated calligraphy images have the features of those conditions.}
         \label{fig:finetune}
     \end{subfigure}

     \caption{Qualitative results of our Calliffusion system.}
     \label{datu}
\end{figure*}

\section{Training}
\subsection{Dataset}
We collected our own dataset by downloading copyright-free Chinese calligraphy images from online. The dataset includes images from 5 scripts, featuring 3975 unique characters and 1431 artists. During the preprocessing stage, we applied a threshold and only retained characters with more than 10 samples, resulting in a reduced dataset of 2025 characters and 1387 artists. 

Additionally, for style transfer with English letters and numbers, we utilized a handwriting dataset from a previous study \cite{de2009character}.
\subsection{Hyperparameters}
We utilized the "diffusers" package\cite{von-platen-etal-2022-diffusers} in Python as the underlying framework for our diffusion models. We configured four blocks in the U-Net architecture with dimensions of 320, 640, 1280, and 1280, each consisting of two layers. We used a Chinese BERT\cite{devlin2018bert} model to obtain cross-attention embeddings with a size of 768 from the input text. The sample size was set to 64, and the batch size was set to 16. We employed the Adam optimizer with a learning rate of $1\times10^{-5}$ and a weight decay of $1\times10^{-6}$. The training was conducted on two NVIDIA A100 40G GPUs for a total of 120 hours.

\begin{table*}
\centering
\resizebox{2\columnwidth}{!}{%
\begin{tabular}{lcccccccccccc}
\hline
                                               & \multicolumn{2}{c}{Regular}                                & \multicolumn{2}{c}{Semi-cursive}                           & \multicolumn{2}{c}{Cursive}                                & \multicolumn{2}{c}{Clerical}                               & \multicolumn{2}{c}{Seal}                                   & \multicolumn{2}{c}{Total}                                  \\
                                               & \multicolumn{1}{l}{Script} & \multicolumn{1}{l}{Character} & \multicolumn{1}{l}{Script} & \multicolumn{1}{l}{Character} & \multicolumn{1}{l}{Script} & \multicolumn{1}{l}{Character} & \multicolumn{1}{l}{Script} & \multicolumn{1}{l}{Character} & \multicolumn{1}{l}{Script} & \multicolumn{1}{l}{Character} & \multicolumn{1}{l}{Script} & \multicolumn{1}{l}{Character} \\ \hline
Real Samples             & 0.91                       & 0.93                          & 0.83                       & 0.81                          & 0.88                       & 0.68                          & 0.96                       & 0.83                          & 0.97                       & 0.81                          & 0.88                       & 0.78                          \\

Generated w/o style condition                   & 0.92                       & \textbf{0.94}                          & \textbf{0.86}                       & 0.89                          & \textbf{0.89} & \textbf{0.72}                          & 0.94                       & 0.87                          & 0.97                       & \textbf{0.80}                 & 0.91                       & 0.84                          \\
Generated w/ style condition & \textbf{0.96}                       & \textbf{0.94} & \textbf{0.86}                       & \textbf{0.95}                          & 0.88                       & 0.64                          & \textbf{0.97}                       & \textbf{0.91}                 & \textbf{0.99}              & 0.79                          & \textbf{0.93}                       & \textbf{0.85}                         
\end{tabular}%
}
\caption{The performance of our generated data in different scripts in accuracy}
\label{table1}
\end{table*}

\section{Calligraphy Generation Examples}
\subsection{Style-free Generation}
Though we used three different conditions to train our models, we do not have to specify all of them during generation. In this section, we show some style-free generation examples by only conditioning on scripts and characters during inference time. 
Figure \ref{datu}(a) and Figure \ref{datu}(b) show generated artworks for two characters, each rendered with 5 different scripts. We see that our models are capable of producing high-quality Chinese calligraphy images, and the controls applied to both character and script are effective.      

\subsection{Stylistic Generation}
We randomly choose 4 characters and render them in an ``unfamiliar'' style, in the sense that the character-style pair never appears in our dataset. In Figure \ref{datu}(c) to Figure \ref{datu}(f), each rendered example is listed together with several real artworks in the corresponding style to showcase the style similarity and consistency. Here, in each of these sub-figures, the first character was generated by our model and the other three are real samples. These examples demonstrate that the control on style is effective, and our later subjective evaluation reveals that even people who know these styles well have difficulty spotting real and generated artworks. 

\subsection{Transfer learning with One-shot Fine-tuning}

During our training, we intentionally leave out a common Chinese character '明' (bright). Later, we handpick two samples, one in regular script by Liu Gongquan and the other in clerical script by Tangtaizong, to fine-tune the model, respectively. After fine-tuning, our model acquired the knowledge of this character and can apply it to other calligraphers' styles. As depicted in the left side of Figure \ref{datu}(i), we generate '明' in regular script with Yan Zhenqin and in clerical script with Caoquanbei.

For digits and English letters, which are certainly not included in our dataset, we pick  '4' and 'a' to conduct fine-tuning. Even with just one-shot, we do not set any specific script or style conditions but only inform the model that the characters are '4' and 'a'. During inference, we incorporate seal script as a condition, and the resulting generated images, as shown on the right-hand side of Figure \ref{datu}(i), exhibit the features of seal script.

\section{Experimental Results}

\subsection{Objective Evaluation}

We used an off-the-shelf pre-trained classifier to recognize the generated images. The classifier is a multitask classifier., whose backbone model is a Res-Net model with two classification embedding layers, one for scripts and one for characters. The generated corpus consists of 2000 images. The first 1000 images are generated by conditioning on the 200 most common characters, each with 5 scripts. We also keep the setting but select 5 famous calligraphers' styles as an extra condition to generate another 1000 images.

The results presented in Table \ref{table1} show that our generated calligraphy is highly similar to real calligraphy. Our generated samples have slightly higher accuracy than the test data (which are real artworks) of the pre-trained classifier. Furthermore, adding style conditions marginally improve the overall accuracies. 

\subsection{Subjective Evaluation}

We designed a survey with three types of questions:

\textbf{Identify the fake artwork:} For each question, we randomly choose a calligrapher's style and select one generated sample produced by our model. The character of the generated sample has never appeared in the collection of the calligrapher's work. Then, we list the generated sample with three genuine artworks composed by the same calligrapher (similar to the layout shown in Figure \ref{datu}(c) to Figure \ref{datu}(f)), asking subjects to identify which of the four choices is generated by an AI model.

\textbf{Identify the real artwork}: The setup is similar to the first type, but the task is to tell the real artwork from the fake ones generated by the model.

\textbf{Identify the script after transfer learning}: For each question of the third type, we use either an English letter or digit generated by our fine-tuned transfer-learning model conditioned with specific scripts and ask subjects whether they can point out the scripts of generated characters. (The generated results are similar to the samples in Figure \ref{datu}(I).)

The survey comprises a total of 10 questions. The first two types of questions contain 4 options each, and a lower accuracy indicates that our generated Chinese calligraphy is highly similar to genuine calligraphy. The third type of question presents 5 options, and a higher accuracy indicates that our generated calligraphy for non-Chinese characters exhibits the characteristics of Chinese calligraphy scripts, making it recognizable to subjects. 

Table \ref{survey} presents the average accuracy and p-value of z score hypothesis testing for each type of question. We collected responses from 150 individuals in China, out of which 87 claimed to have practiced Chinese calligraphy or know the scripts and style used in the survey. The null hypothesis in this study is that the accuracy for each question is equal to random guessing (25\% for questions with 4 options and 20\% for questions with 5 options). 

For the first two types of questions, the accuracy for individuals with previous knowledge of Chinese calligraphy is slightly higher than random guessing, whereas, for those who are unfamiliar with calligraphy, the accuracy is slightly lower. The p-values show that the results are not significantly different from random guess. In contrast, the third type of question revealed that around 70\% of the subjects were able to identify the calligraphy script characteristics in our generated non-Chinese symbols, and the p-value indicates that this result is significant at $p < 0.001$.

\begin{table}[]
\begin{tabular}{llll}
              & Know Calli                    & Don't know                    & \multicolumn{1}{c}{Total}     \\ \hline
No.  & \multicolumn{1}{c}{87}        & \multicolumn{1}{c}{63}        & \multicolumn{1}{c}{150}       \\ \hline
Q.     & \multicolumn{1}{c}{Acc(P-Val)} & \multicolumn{1}{c}{Acc(P-Val)} & \multicolumn{1}{c}{Acc(P-Val)} \\ \hline
1$\downarrow$        & 0.275(0.296)                   & 0.245(0.853)                   & 0.263(0.459)                  \\
2$\downarrow$        & 0.286(0.293)                   & 0.246(0.917)                   & 0.269(0.458)                   \\
3$\uparrow$        & 0.796(***)                     & 0.579(***)                     & 0.706(***)                    
\end{tabular}
\caption{The accuracy and p-value of each type of question in our survey.}
\label{survey}
\end{table}

\section{Limitation}
\begin{figure}[!htb]
     \centering
     \begin{subfigure}[b]{0.2\textwidth}
         \centering
         \includegraphics[width=\textwidth]{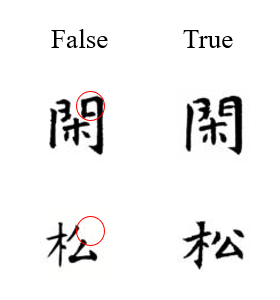}
         \caption{Generated failures with missing strokes.}
         \label{fig:miss}
     \end{subfigure}
     \hfill
     \begin{subfigure}[b]{0.2\textwidth}
         \centering
         \includegraphics[width=\textwidth]{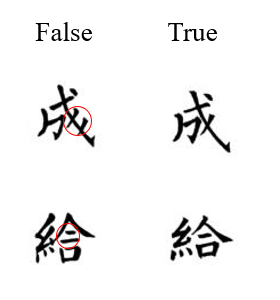}
         \caption{Generated failures with unnecessary extra strokes.}
         \label{fig:add}
     \end{subfigure} 
     \caption{Comparison of unsuccessful and successful results of our Calliffusion system.}
    \label{fail}
\end{figure}

In this section, we presented examples of unsuccessful generated outcomes that could potentially pass an objective classification assessment but can be easily identified by humans familiar with the Chinese language. These failures can be categorized into two primary types: the missing of certain strokes, shown in Figure \ref{fail}(a), or the addition of unnecessary strokes, shown in Figure \ref{fail}(b). According to our experiment, we discovered that increasing the amount of training data and conducting more training epochs can lead to a reduction in the number of generated failures.

\section{Conclusion}

In this paper, we introduce a conditional diffusion model for generating Chinese calligraphy. We demonstrate that our model can produce high-quality calligraphy by conditioning it with various combinations of characters, scripts, and styles. Additionally, we can generate previously unseen Chinese characters or even non-Chinese symbols using a one-shot transfer learning with LoRA. 
The artworks produced by our model undergo assessment through both objective and subjective evaluations. The objective evaluation demonstrates that our generated calligraphy exhibits exceptional accuracy when classified by a pre-trained classifier. The subjective evaluation indicates that when our generated samples are compared with authentic calligraphy, discerning between the two becomes exceedingly challenging for human observers.

Moving forward, our aim is to delve into the realm of few-shot style transfer learning for novel styles and scripts. Currently, we have the capability to perform style transfer for new characters or symbols using a one-shot approach. The future plan is to discover an effective method to learn new scripts beyond the five conventional Chinese calligraphy scripts or acquire new styles from the handwriting of any individual, which could make our model become even more valuable and versatile.

\bibliographystyle{iccc}
\bibliography{iccc}

\begin{thebibliography}{}

\bibitem[\protect\citeauthoryear{De~Campos \bgroup et al.\egroup
  }{2009}]{de2009character}
De~Campos, T.~E.; Babu, B.~R.; Varma, M.; et~al.
\newblock 2009.
\newblock Character recognition in natural images.
\newblock {\em VISAPP (2)} 7(2).

\bibitem[\protect\citeauthoryear{Devlin \bgroup et al.\egroup
  }{2018}]{devlin2018bert}
Devlin, J.; Chang, M.-W.; Lee, K.; and Toutanova, K.
\newblock 2018.
\newblock Bert: Pre-training of deep bidirectional transformers for language
  understanding.
\newblock {\em arXiv preprint arXiv:1810.04805}.

\bibitem[\protect\citeauthoryear{Goodfellow \bgroup et al.\egroup
  }{2014}]{NIPS2014_5ca3e9b1}
Goodfellow, I.; Pouget-Abadie, J.; Mirza, M.; Xu, B.; Warde-Farley, D.; Ozair,
  S.; Courville, A.; and Bengio, Y.
\newblock 2014.
\newblock Generative adversarial nets.
\newblock In Ghahramani, Z.; Welling, M.; Cortes, C.; Lawrence, N.; and
  Weinberger, K., eds., {\em Advances in Neural Information Processing
  Systems}, volume~27.
\newblock Curran Associates, Inc.

\bibitem[\protect\citeauthoryear{Ho, Jain, and Abbeel}{2020}]{ho2020denoising}
Ho, J.; Jain, A.; and Abbeel, P.
\newblock 2020.
\newblock Denoising diffusion probabilistic models.
\newblock {\em Advances in Neural Information Processing Systems}
  33:6840--6851.

\bibitem[\protect\citeauthoryear{Hu \bgroup et al.\egroup }{2021}]{hu2021lora}
Hu, E.~J.; Shen, Y.; Wallis, P.; Allen-Zhu, Z.; Li, Y.; Wang, S.; Wang, L.; and
  Chen, W.
\newblock 2021.
\newblock Lora: Low-rank adaptation of large language models.
\newblock {\em arXiv preprint arXiv:2106.09685}.

\bibitem[\protect\citeauthoryear{Ronneberger, Fischer, and
  Brox}{2015}]{ronneberger2015u}
Ronneberger, O.; Fischer, P.; and Brox, T.
\newblock 2015.
\newblock U-net: Convolutional networks for biomedical image segmentation.
\newblock In {\em Medical Image Computing and Computer-Assisted
  Intervention--MICCAI 2015: 18th International Conference, Munich, Germany,
  October 5-9, 2015, Proceedings, Part III 18},  234--241.
\newblock Springer.

\bibitem[\protect\citeauthoryear{Tian}{2017}]{zi2zi}
Tian, Y.
\newblock 2017.
\newblock zi2zi: Master chinese calligraphy with conditional adversarial
  networks.

\bibitem[\protect\citeauthoryear{von Platen \bgroup et al.\egroup
  }{2022}]{von-platen-etal-2022-diffusers}
von Platen, P.; Patil, S.; Lozhkov, A.; Cuenca, P.; Lambert, N.; Rasul, K.;
  Davaadorj, M.; and Wolf, T.
\newblock 2022.
\newblock Diffusers: State-of-the-art diffusion models.

\bibitem[\protect\citeauthoryear{Wen \bgroup et al.\egroup
  }{2021}]{wen2021zigan}
Wen, Q.; Li, S.; Han, B.; and Yuan, Y.
\newblock 2021.
\newblock Zigan: Fine-grained chinese calligraphy font generation via a
  few-shot style transfer approach.
\newblock In {\em Proceedings of the 29th ACM International Conference on
  Multimedia},  621--629.

\bibitem[\protect\citeauthoryear{Wu, Yang, and Hsu}{2020}]{wu2020calligan}
Wu, S.-J.; Yang, C.-Y.; and Hsu, J. Y.-j.
\newblock 2020.
\newblock Calligan: Style and structure-aware chinese calligraphy character
  generator.
\newblock {\em arXiv preprint arXiv:2005.12500}.

\bibitem[\protect\citeauthoryear{Zhou \bgroup et al.\egroup
  }{2021}]{zhou2021end}
Zhou, P.; Zhao, Z.; Zhang, K.; Li, C.; and Wang, C.
\newblock 2021.
\newblock An end-to-end model for chinese calligraphy generation.
\newblock {\em Multimedia Tools and Applications} 80:6737--6754.

\end{thebibliography}

\end{CJK}
\end{document}